\documentclass[runningheads]{llncs}
\usepackage{graphicx}
\usepackage{color}
\usepackage[left=2.54cm,top=2.54cm,right=2.54cm,bottom=2.54cm]{geometry}
\usepackage{float}
\usepackage[caption=false]{subfig}
\usepackage[spanish]{babel}
\usepackage{bigstrut}
\usepackage{array}
\usepackage{multirow}
\usepackage{color}
\usepackage[dvipsnames]{xcolor}
\usepackage{xcolor, colortbl}
\usepackage[sorting=none]{biblatex}
\usepackage{tikz}
\usetikzlibrary{arrows,positioning,shapes.geometric}
\usepackage{csquotes}
\DeclareUnicodeCharacter{0301}{\'}
\DeclareUnicodeCharacter{0308}{\" }

\begin{document}
\title{Estimación de áreas de cultivo mediante Deep Learning y programación convencional}
\titlerunning{Preprint}  

\author{Javier Caicedo\inst{1} \and Pamela Acosta\inst{1} \and Romel Pozo\inst{1} \and
Henry Guilcapi\inst{1} \and Christian Mejia-Escobar\inst{1}\thanks{Email de correspondencia: cimejia@uce.edu.ec}}
\institute{Universidad Central del Ecuador, Facultad de Ingeniería en Geología, Minas, Petróleos y Ambiental, Quito-Ecuador}

\maketitle              

\begin{abstract}
La Inteligencia Artificial ha permitido la implementación de soluciones más precisas y eficientes para problemáticas en diversas áreas. En el sector agrícola, una de las principales necesidades es conocer en todo momento la extensión de terreno ocupada o no por los cultivos con el fin de mejorar la producción y la rentabilidad. Los métodos tradicionales de cálculo demandan la obtención de datos de manera manual y presencial en campo, ocasionando altos costes de mano de obra, tiempos de ejecución, e imprecisión en los resultados. El presente trabajo propone un nuevo método basado en técnicas de Deep Learning complementadas con programación convencional para la determinación del área de zonas pobladas y despobladas de cultivos. Hemos considerado como caso de estudio una de las empresas más reconocidas en la siembra y cosecha de caña de azúcar en el Ecuador. La estrategia combina una Red Neuronal Adversaria Generativa (GAN) que es entrenada sobre un dataset de fotografías aéreas de paisajes naturales y urbanos para mejorar la resolución de imágenes; una Red Neuronal Convolucional (CNN) entrenada sobre un dataset de fotografías aéreas de parcelas de caña de azúcar para distinguir zonas pobladas o despobladas de cultivo; y un módulo de procesamiento estándar de imágenes para el cálculo de áreas de manera porcentual. Los experimentos realizados demuestran una mejora significativa de la calidad de las fotografías aéreas así como una diferenciación notable entre las zonas pobladas y despobladas de cultivo, consecuentemente, un resultado más preciso de las áreas cultivadas y no cultivadas. El método propuesto puede ser extendido a la detección de posibles plagas, zonas de vegetación de maleza, el desarrollo dinámico de los cultivos, y control de calidad tanto cualitativo como cuantitativo.


\end{abstract}

\section{Introducción}

La fotografía aérea es una fuente de información fundamental para las Ciencias de la Tierra. El proceso de examinarlas para identificar objetos o ciertas condiciones se denomina \textit{fotointerpretación} \cite{fernandez2000introduccion}. Por ejemplo, dentro del área geológica y agrícola permite delimitar diversos parámetros tales como topografía, rasgos hidrográficos, litología en función de texturas y características agrícolas como superficies de cultivos, delimitación y análisis focalizado en áreas de cultivos, entre otras, que desde superficie no se pueden apreciar de forma óptima \cite{matarredona1985aplicacion}. En sus inicios, la fotointerpretación debía ser realizada de manera manual por operadores especializados, significando mayor gasto de recursos y altos tiempos de espera para las compañías. Más aún, cuando dichas imágenes no eran capturadas en una resolución adecuada, el trabajo del interpretador muchas veces no era posible de realizar o podía tomar mucho más tiempo de lo esperado. Un avance importante en este ámbito se logra con el uso de los \textit{GIS} (Sistemas de Información Geográfica)\cite{blanco2013integracion}, a través del procesamiento de las fotografías aéreas aplicando operaciones, tanto cualitativas como cuantitativas, como por ejemplo, medición de áreas, cálculos topográficos, descripciones geológicas como texturas asociadas a tipos de roca o suelos, fenómenos de remoción en masa, delimitación de formaciones a través del reconocimiento en la disposición de las capas, etc., para una toma de decisiones adecuada. Sin embargo, aún se requiere de un operador que controle dichas herramientas de interpretación, necesitando de una mayor cantidad de tiempo, esfuerzo y habilidad por parte del ser humano.

Hoy en día, las técnicas de Inteligencia Artificial (IA), en particular el Deep Learning se ha convertido en una herramienta exitosa para el procesamiento automático de imágenes de diversa índole, así como para el reconocimiento y delimitación de objetos de manera automatizada. Es un concepto que surge de la idea de imitar el cerebro y el sistema visual a partir del uso de hardware y software, para crear una inteligencia artificial pura, utilizando una capacidad de abstracción jerárquica, es decir, una representación de los datos de entrada en varios “niveles” \cite{restrepo2015aplicacion}.


En el presente trabajo, proponemos la implementación de un sistema basado en Deep Learning, para el procesamiento automático de fotografías aéreas, que permita mejorar la resolución de ortofotografías capturadas en baja resolución para posteriormente realizar la identificación y cálculo de superficies relacionadas con zonas pobladas y despobladas en cultivos de caña de una manera automática. Hemos considerado el caso de la compañía Azucarera Valdez, la cual está ubicada en el cantón Milagro de la provincia del Guayas (Ecuador), con una producción de azúcar ininterrumpida durante 138 años, posicionándose como una de las industrias más grandes del país en el ámbito alimenticio. Actualmente se busca automatizar procedimientos manuales en campo, como son las actividades de muestreo para despoblación, plagas, malezas entre otras, con la finalidad de optimizar el flujo de trabajo en todas las áreas involucradas. Es de vital importancia para la empresa, conocer el desarrollo de los cultivos de caña a través de fotografías aéreas, con el propósito de llevar un control cuantitativo de las áreas cultivadas y no cultivadas.

Nuestra estrategia de solución es combinar las \textit{Redes Neuronales Adversarias Generativas} (GAN), las \textit{Redes Neuronales Convolucionales} (CNN) y la programación tradicional. Las GANs consisten en dos modelos, el generador y el discriminador, entrenados simultáneamente para desafiarse uno al otro, lo que explica el término antagónicas elegido por los autores para darle identidad a este novedoso método \cite{calcagni2020redes}. La generadora es entrenada para crear datos falsos lo más parecidos posibles a los ejemplos reales de un determinado conjunto de entrenamiento. Por otro lado, la discriminadora es entrenada para ser capaz de discernir los datos falsos producidos por la generadora de aquellos que corresponden al conjunto de entrenamiento (los ejemplos reales). Dicha estructura es útil para el aumento de la resolución de las fotografías aéreas que requieran una optimización mediante operaciones de píxel con el objetivo de reducir el tamaño y reorganizar la información permitiendo así transponer la misma a través de la manipulación de tensores e ir mejorando la imagen\cite{loncaric2021integracion}. Estas imágenes serán el insumo para una CNN, la cual funciona a través de capas de neuronas que se conforman de píxeles representados por números, y que se operan matemáticamente a través de la aplicación de filtros o kernels para dar como resultado un mapa de características de la imagen original \cite{picazo2018redes}. Estas características son pasadas a un clasificador que se encargará del reconocimiento y categorización de superficies en las fotografías determinándolas como pobladas (referido a toda la zona cultivada de caña) o despobladas (zonas sin cultivo de caña). Por el enfoque de la problemática se tomará en cuenta únicamente las zonas que no presentan cultivo dentro del área delimitada y se discriminará carreteras y drenajes que estén presentes en las fotografías mediante una delimitación previa en la planificación del vuelo donde solamente se considerará el área cultivada. Finalmente, se calculará de forma porcentual las superficies que presenten despoblación en los cultivos de caña que será el dato de interés, en consecuencia también se obtendrá el dato del área que se encuentra cultivada, a través de la utilización de bibliotecas de programación tradicional.

De esta manera, es posible optimizar los tiempos, costos, logística e insumos de la compañía, cuya necesidad de conocer parámetros de sus cultivos se ha visto encarecida por la obtención de datos en forma tradicional, debido al costo de mano de obra y con el riesgo de un alto porcentaje de error por la inaccesibilidad en terrenos. Adicionalmente, el mejoramiento y análisis automático de fotografías aéreas tiene una amplia gama de aplicabilidad, siendo factible transponer los resultados de este estudio hacia una distinta problemática que puede o no estar en la misma área de conocimiento como identificación de malezas, zonas con problemas de floración en caña, delimitación de control de rebrotes en diferentes cultivos. Concretamente, proporcionamos las siguientes contribuciones:
 
\begin{itemize}
    \item Un dataset de 650 fotografías aéreas de alta y baja resolución obtenidas del portal SIGTIERRAS y que muestran entornos urbanos, rurales, y naturales para el entrenamiento de Real-ESRGAN.
    \item Un dataset de 1600 imágenes restringidas de los predios de la compañía azucarera Valdez que muestran áreas de cultivos con zonas pobladas y despobladas.  
    \item Una metodología que combina las GANs, CNNs y la programación tradicional para procesar imágenes aéreas con el fin de automatizar procedimientos de interpretación de información morfométrica de cultivos de caña y que puede ser adaptada para problemas similares.
    \item La implementación práctica de modelos de súper-resolución, categorización y tratamiento de imágenes, aprovechando la técnica del \textit{Transfer Learning} y librerías tradicionales de procesamiento de imágenes.  
\end{itemize}

Los productos mencionados estarán disponibles en la Escuela de Geología de la Universidad Central del Ecuador para fin\textbf{}es educativos, ya que por motivos de confidencialidad relacionados con políticas de la Compañía Valdez, los datos utilizados en el presente estudio no podrán ser publicados en un repositorio digital, sin embargo, podrían ser solicitados a cualquiera de los autores.

El resto del documento se estructura de la siguiente manera: en la sección 2 se citan trabajos relacionados; la sección 3 describe detalladamente la metodología empleada; la sección 4 explica los experimentos realizados y los resultados que se obtuvieron para cada uno de los modelos trabajados; la sección 5 presenta la discusión de los mismos; y finalmente, la sección 6 incluye las conclusiones del estudio y algunas líneas de trabajo futuro.

\section{Trabajos relacionados}
\label{related-work}

El uso de CNNs y GANs se ha ido ampliando durante los últimos años, sin embargo, no conocemos de un trabajo previo que combine ambos tipos de redes neuronales artificiales para tratar el problema específico de cálculo de zonas pobladas y despobladas de cultivo a partir de fotografías aéreas \cite{ospina2021aplicacion}. Existen ciertos estudios que aprovechan por separado el uso de CNNs y GANs para abordar temáticas relacionadas con fotografías aéreas y geociencias.

\subsection{Uso de GAN's para súper-resolución}
En cuanto a la súper-resolución de imágenes podemos citar a Clabaut. É, et al., 2021 \cite{clabaut2021model}, con su estudio que se titula “Model Specialization for the Use of ESRGAN on Satellite and Airborne Imagery”, donde utilizan el modelo previo a Real-ESRGAN, desarrollado por Xintao Wang. Aquí se propone evaluar el entrenamiento de una GAN para mejorar la resolución de imágenes satelitales. Los autores utilizan varios datasets especializados en imágenes aéreas de varias categorías (urbanas, rurales, marítimas) donde se hace una comparación de los resultados de un entrenamiento al realizar una mezcla de los datasets y al entrenar al modelo de manera independiente por cada categoría. Los resultados muestran un mejor funcionamiento al entrenar el modelo para una categoría específica y lo evalúan a través del ruido generado durante el proceso de downscale.

\subsection{Uso de CNN's para reconocimiento}
Aldás. R, et al., 2022 \cite{aldas2022delimitacion}, en su trabajo denominado “Delimitación automática de ceniza volcánica en imágenes satélites mediante Deep Learning", implementan una solución basada en aprendizaje profundo que permite segmentar emisiones de ceniza en imágenes satélites mediante Redes Neuronales Convolucionales. A través de la elaboración de un extenso dataset de imágenes, entrenaron un modelo para la predicción de la delimitación del esparcimiento de ceniza volcánica de nuevas imágenes satélites. De esta forma, generan un aporte en cuanto a la gestión de las zonas afectadas por el fenómeno de caída de ceniza y su impacto en sectores estratégicos como la agricultura, ganadería y la salud. 

\subsection{Combinación de GAN's y CNN's}
Maayan Frid-Adar, et al., 2018 \cite{frid2018gan}, con su estudio “Aumento de imágenes médicas sintéticas basadas en GAN para un mayor rendimiento de CNN en la clasificación de lesiones hepáticas”, presenta métodos para generar imágenes médicas sintéticas utilizando redes generativas adversarias (GAN) y así obtener imágenes de lesiones hepáticas de alta calidad. Dichas imágenes generadas se pueden usar para el aumento de datos sintéticos y mejorar el rendimiento de una red CNN utilizada en la clasificación de imágenes médicas usando un conjunto limitado de 182 tomografías (53 quistes, 64 metástasis y 65 hemangiomas). Entrenando a la red CNN se obtuvo un rendimiento de sensibilidad de 78.6\% con una especificidad de 88.4\% generando un esquema para la clasificación de lesiones hepáticas. Este enfoque puede ayudar a respaldar los esfuerzos de los radiólogos para mejorar el diagnóstico sobre lesiones hepáticas \cite{perez2011mejora}. 

Yang Li y Xuewei Chao, 2020 \cite{li2020ann}, con su trabajo “Clasificación continua basada en ANN en agricultura” realizan una clasificación oportuna de enfermedades de plantas a través de imágenes. Los autores proponen un método de clasificación basado en ANN a través del almacenamiento y recuperación de memoria, con dos claras ventajas: pocos datos y alta flexibilidad. Este modelo combina una CNN y una GAN. La red CNN requiere pocos datos sin procesar para lograr un buen rendimiento, para una tarea de clasificación de plantas y la parte GAN se usa para extraer información de tareas antiguas y generar imágenes abstractas como memoria para la tarea futura. La red ANN se puede utilizar para tareas nuevas, con buen desempeño, debido a su capacidad de acumular conocimiento. La protección de las plantas se logró mediante la identificación automática de enfermedades y plagas \cite{piscoya2019sistema}. 

A diferencia de los trabajos anteriores, nuestro enfoque está dirigido hacia una problemática en el campo de la denominada \textit{agricultura de precisión}, donde la particularidad radica en que en este caso nosotros diseñamos un flujo de trabajo robusto, que combina técnicas de aprendizaje automático y programación tradicional para lograr enfrentar el problema en cuestión, mejorando tiempos, reducir costos y riesgos y obtener una mayor precisión en las mediciones son varias de las ventajas que podremos materializar con este estudio. Adicionalmente consideramos que los modelos y las arquitecturas utilizadas ofrecen la versatilidad necesaria para permitir adaptar a este flujo de trabajo a temáticas relacionadas. 

\section{Metodología}
\label{Metodología}
\begin{figure}[H]
 \centering
 \includegraphics[width=0.9\linewidth]{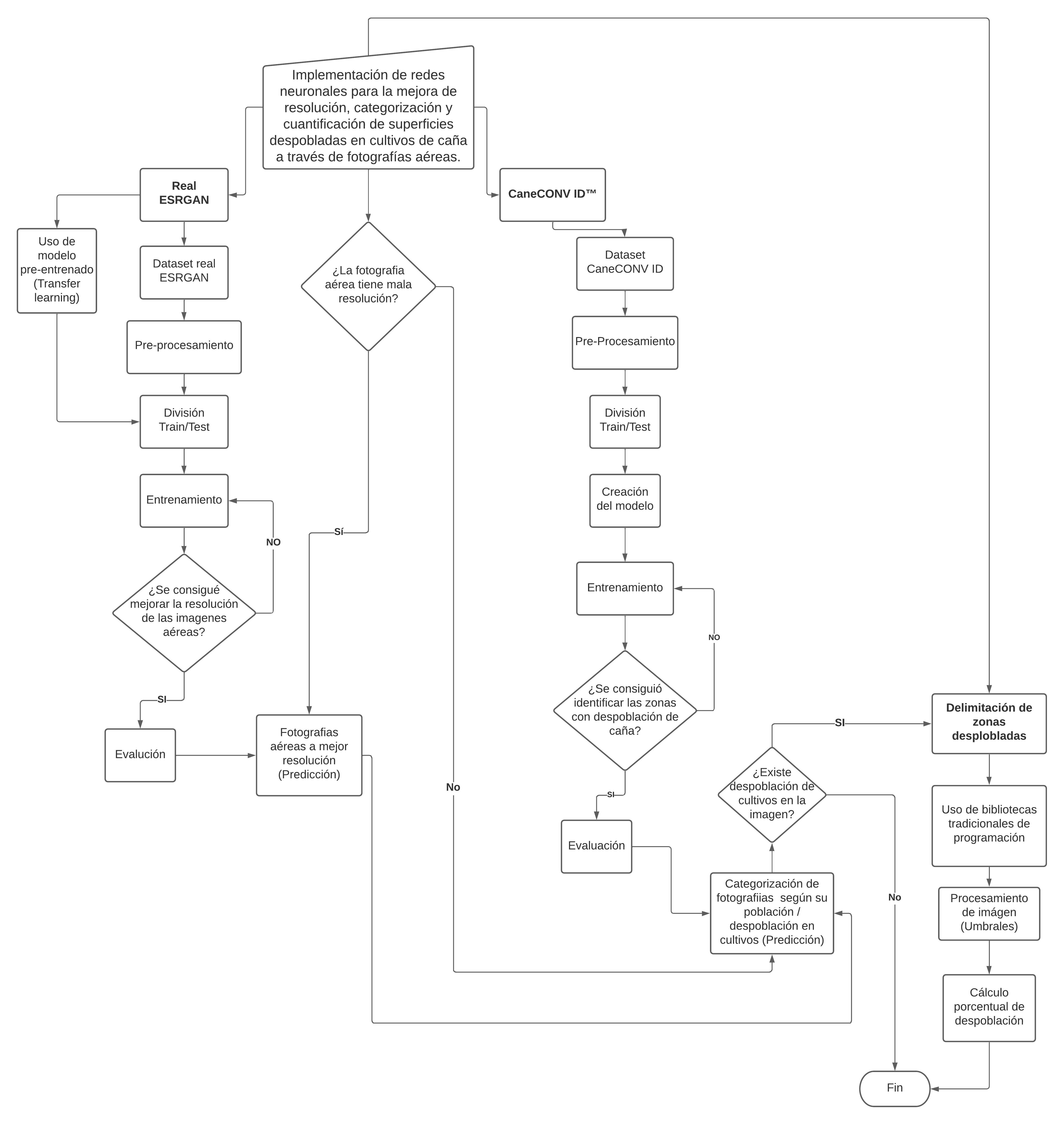}
 \caption{\label{fig:metodologia} Metodología de trabajo. Elaborado por: Autores}
\end{figure}

Nuestro objetivo es el cálculo automático del porcentaje de zonas cultivadas y no cultivadas de caña de azúcar en los lotes de plantación de la empresa Azucarera Valdez. Para tal fin, combinamos técnicas de Inteligencia Artificial y la programación tradicional, conformando una solución compuesta de 3 módulos. Primero, utilizamos una red de tipo GAN que será entrenada para aumentar la resolución de las fotografías aéreas de los lotes, lo cual contribuye para una mejor categorización y mayor precisión en el cálculo de extensiones de cultivo. Luego, aprovechamos una red convolucional convencional entrenada para distinguir la presencia de población o despoblación de cultivos de caña. Finalmente, en el caso de detectar la existencia de áreas con despoblación de cultivos, se procede a realizar la delimitación porcentual de dicha extensión mediante librerías de procesamiento de imágenes y programación tradicional, en caso de no detectar la existencia de despoblación de cultivos no es posible un cálculo referencial de dicha área, debido a que su extensión pertenecerá exclusivamente a áreas con población de cultivos. El flujo de trabajo está representado en la Figura \ref{fig:metodologia} y, seguidamente, describimos en detalle cada una de las etapas consideradas. 

\subsection{Súper-resolución con Real-ESRGAN}
\subsubsection{Dataset}
La materia prima para el aprendizaje automático de tipo supervisado es el conjunto de datos, cada uno asociado con su respectiva respuesta. En nuestro caso, corresponden a imágenes de resolución baja y su respectivo par de alta resolución. Así, fue necesaria una recolección de imágenes que permitieran tener una gama de categorías adecuada para entrenar al modelo en función de nuestro objetivo. Todas estas imágenes fueron obtenidas a través del portal del Sistema Nacional de Información de Tierras Rurales e Infraestructura Tecnológica (SIGTIERRAS)\footnote{\url{www.sigtierras.gob.ec}}, donde se requiere crear un usuario para solicitar fotografías aéreas disponibles de forma gratuita y que cubren la mayor parte del territorio nacional. Se eligieron zonas para ortofotografía donde se pudieran observar espacios urbanos y rurales o áreas naturales. Se consideraron ortofotografías de ciudades del Ecuador como Quito, Guayaquil, Manta y Salinas dentro del apartado urbano, con un total de 250 fotografías y se complementó con 400 fotografías de ciudades como El Puyo, Milagro, Tena y Sucumbíos para el apartado rural y entornos naturales.

\begin{figure}
 \centering
    \includegraphics[scale=.3]{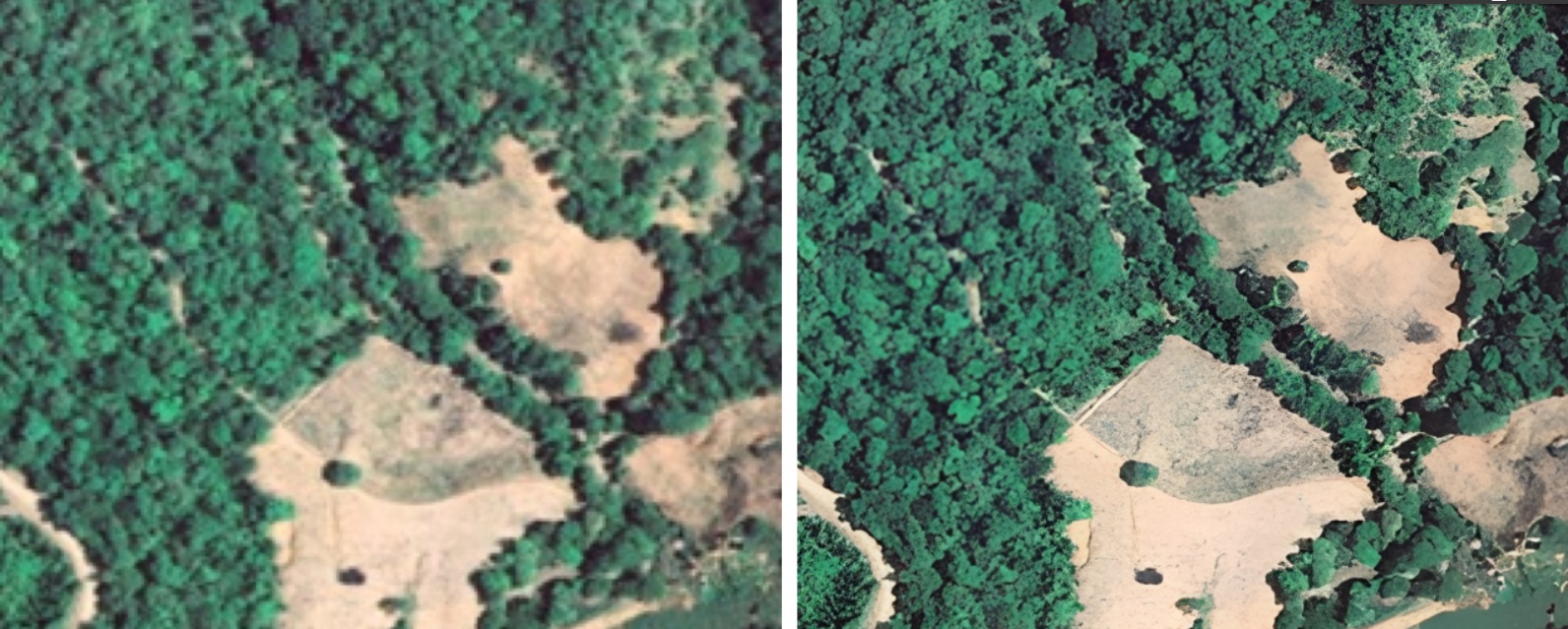}
    \paragraph{}
    \includegraphics[scale=0.297]{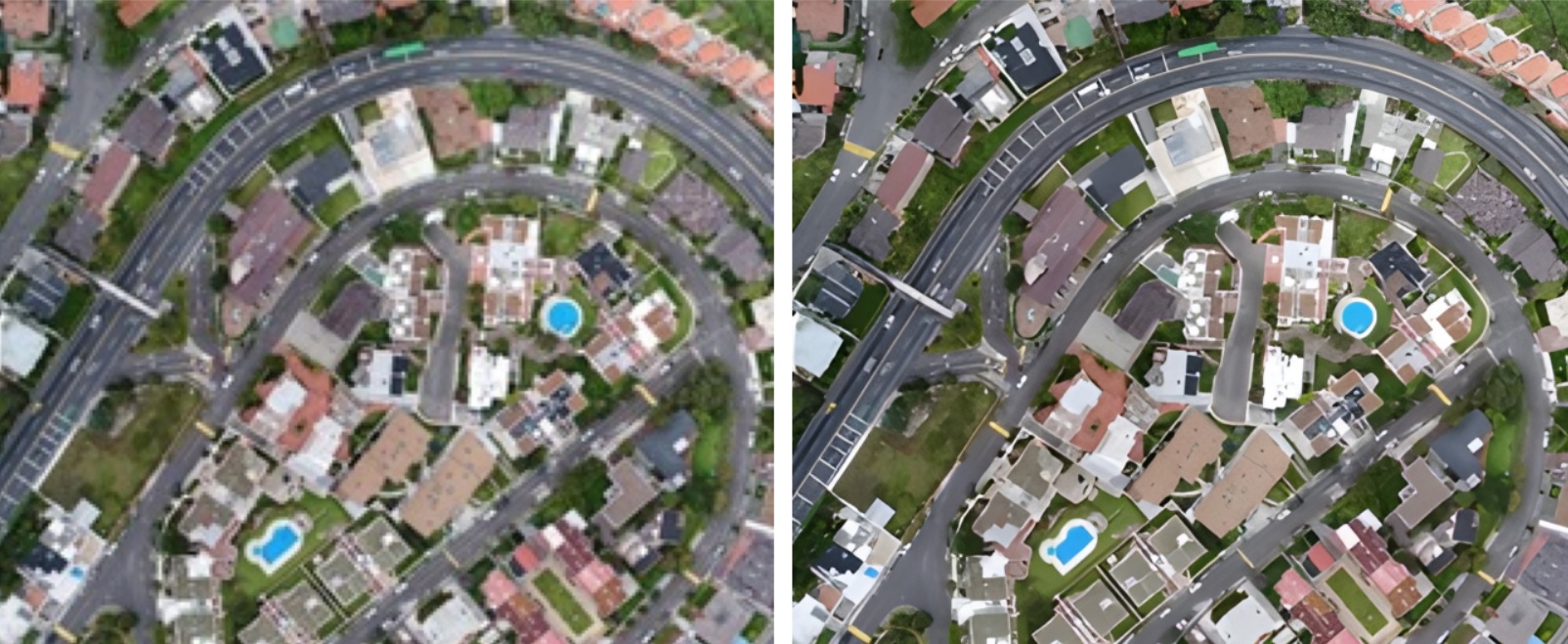}
    \caption{\label{fig:muestradataset} Ejemplos de imágenes aéreas del dataset: (sup) Ejemplo de la categoría rural o área natural, (inf) Ejemplo de la categoría urbana. Fuente: Portal SIGTIERRAS.}
\end{figure}

El portal de SIGTIERRAS ofrece fotografías a una resolución de 5 metros en formato JP2, el cual permite guardar datos de georreferenciación. Se crearon dos directorios para la construcción del dataset: ''URBANO'' y ''RURAL'', con un total de 650 imágenes con dimensiones de (4647x4617), divididas en dos partes, de tal manera que para la zona urbana se utilizaron 250 imágenes y 400 imágenes para la zona rural, con un peso total de 24 GB, sin embargo, la denominación de las imágenes está de acuerdo al rollo y a la zona de captura correspondiente a la hoja topográfica de las zonas de interés. La Figura \ref{fig:muestradataset} muestra ejemplos de fotografías de baja resolución con su par de alta resolución utilizados en el dataset para el entrenamiento del modelo Real-ESRGAN; además, como se puede apreciar en este ejemplo, las imágenes de la izquierda se encuentran pixeladas, con contraste de colores opacos y los límites de los objetos son difusos tanto para la categoría urbana como rural, mientras que en las imágenes de la derecha, las cuales han sido entrenadas con el modelo Real-ESRGAN, poseen una calidad superior, con límites bien definidos y mejor contraste de colores para ambas categorías \cite{britez2021exploratory}.

\subsubsection{Pre-procesamiento}

Las imágenes descargadas y organizadas en las dos carpetas mencionadas, son convertidas del formato original JP2 de tipo georreferenciado a un formato simple JPG, ampliamente utilizado para el manejo de fotografías en todo tipo de dispositivos, además caracterizado por una buena compresión. Este procedimiento se realizó en lotes de 20 fotografías mediante la utilización del convertidor online \textit{convertio}\footnote{\url{https://convertio.co/es/jp2-jpg/}}. Para la denominación de las imágenes, se mantuvo el nombre por defecto de la descarga desde el portal SIGTIERRAS.

\subsubsection{División en Train/Test}

Los algoritmos de aprendizaje automático aprenden de los datos con los que los entrenamos. A partir de ellos, intentan encontrar o inferir el patrón que les permita predecir el resultado para un nuevo caso. Pero, para poder calibrar si un modelo funciona, necesitaremos probarlo con un conjunto de datos diferente. Por ello, en todo proceso de aprendizaje automático, los datos de trabajo se dividen en dos partes: una parte \textit{Train} o entrenamiento, que corresponderá a la mayor parte de nuestro dataset y que usaremos para entrenar nuestro modelo y una parte \textit{Test}, de menor tamaño, sobre la que evaluaremos nuestro modelo entrenado \cite{badenes2022conceptualizacion}. 
Para nuestro caso, el proceso de división se realizó automáticamente utilizando función \textit{train\_test\_split()} de la librería \textit{sklearn.modelselection}, especificando dos subconjuntos: el de entrenamiento que contiene el 80\% de los datos y el de prueba con el 20\% de los datos restantes. Estas proporciones de división son usualmente recomendadas para asegurar una mayor densidad de datos en el conjunto de entrenamiento, y menos en el conjunto de prueba. Es importante mencionar que la subdivisión fue realizada con un carácter aleatorio para garantizar la mayor variabilidad en ambos subconjuntos. La Tabla \ref{tab:ComposicionDatasetGAN} muestra la distribución de los datos en la subdivisión.  

\begin{table}[H]
\centering
\begin{tabular}{|c|c|c|}
\hline
\textbf{Conjunto} & \textbf{Número de Imágenes} & \textbf{Porcentaje} \\ \hline
Train     & 520                         & 80\%                \\ \hline
Test      & 130                         & 20\%                \\ \hline
Total  & 650                         & 100\%               \\ \hline
\end{tabular}
\vspace{0.2cm}
\caption{Subdivisión del dataset.}
\label{tab:ComposicionDatasetGAN}
\end{table}

\subsubsection{Modelo}

Hemos aprovechado la técnica de \textit{Transfer Learning}, misma que se basa en una idea simple, la de re-utilizar los conocimientos adquiridos por otras configuraciones (fuente), para resolver un problema en particular (objetivo) \cite{Datascientest}. En otras palabras, nos induce a la adaptación de modelos ya creados, a nuestro convenir. Las ventajas de la aplicación de esta técnica son variadas, pero destacan la optimización del tiempo y los buenos resultados que proporciona, producto del enfoque que previamente la red ya ha recibido.

Para este caso, hemos utilizado el modelo GAN para súper-resolución de imágenes desarrollado por (Wang. X, et al., 2021)\footnote{\url{https://github.com/xinntao/Real-ESRGAN}}. La arquitectura del modelo se ha mantenido intacta, únicamente fue necesario someterla a un nuevo entrenamiento sobre nuestro dataset. En este sentido, es importante dilucidar la arquitectura con la que Real-ESRGAN trabaja. Utiliza una arquitectura que busca optimizar la estabilidad durante el entrenamiento, al mismo tiempo que trata de eliminar o mitigar el blur, el ruido, el re-dimensionamiento y la compresión \cite{heras2017clasificador}, los cuales se generan como el resultado de una mala captura de las imágenes en el caso de fotografías, o por el trayecto que una imagen recorre cuando es cargada a la Web. Todo esto reduce la calidad de la imagen. 

\begin{figure}[H]
\centering
\includegraphics[width=0.99\linewidth]{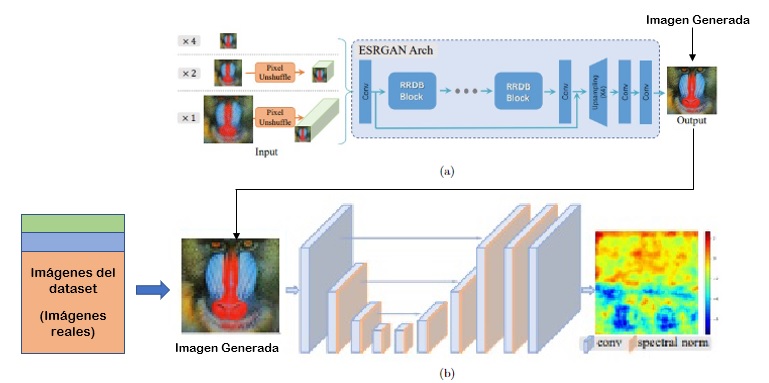}
\label{fig:rrbd}
\caption{Arquitectura del modelo GAN: (a) Esquema de la red generadora (RRDB); (b) Red discriminadora (U-net). Modificado de: Wang. X, et al., 2021 \cite{wang2021real}.}
\label{fig:GANmodel1}
\end{figure}

\begin{figure}[H]
\centering
\includegraphics[width=0.9\linewidth]{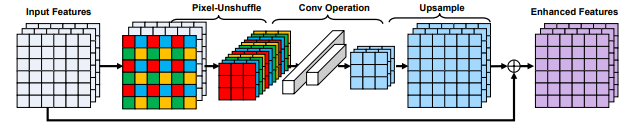}
\label{fig:unshuffle}
\caption{Esquema representativo de la operación de pixel-Unshuffle. Tomado de: (Sun. B et al. 2019) \cite{sun2022hybrid}}
\label{fig:GANmodel2}
\end{figure}

\begin{figure}
 \centering
    \includegraphics[scale=1]{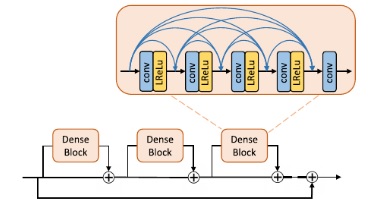}
    \caption{\label{fig:rrdb} Esquema de la estructura de los RRDB. Tomado de: Wang. H et al. 2019} \cite{wang2019deformable}.
\end{figure}

La arquitectura de este modelo se basa en un diseño tradicional de las GANs, donde intervienen una red generadora y una red discriminadora (Figura \ref{fig:GANmodel1}). La red generadora funciona de la siguiente manera:
\begin{itemize}
    \item Antes del input de la imagen al modelo es necesaria una operación de \textit{Pixel-Unshuffle} cuyo objetivo es reducir el tamaño espacial y reorganizar la información a la dimensión del canal (Figura \ref{fig:unshuffle}). Esta operación básicamente permite transponer la información desde una dimensión espacial a una dimensión de profundidad a través de la manipulación de tensores \cite{wang2021real}. La dimensión de profundidad representa el resultado de la descomposición y combinación de píxeles expresado a través de la creación de tensores. Esta operación divide una característica en varias subcaracterísticas extraídas de la imagen original. En el ejemplo se usan cuatro colores diferentes para representar subcaracterísticas de la imagen de entrada. Estas subcaracterísticas (que en realidad son píxeles) contienen la información completa de las características originales de la imagen, pero con menor resolución \cite{sarango2022redes}. Por lo tanto, se utiliza este procedimiento para evitar pérdida de información mientras se reduce el tamaño de la imagen de entrada. Una vez obtenidas todas los tensores es necesario convolucionarlos con kernels para obtener un pre-escalado de la imagen de entrada.
    \item Seguidamente, la imagen resultante del pixel unshuffle ingresa a la red generadora, misma que se cataloga como una red profunda conformada por varios \textit{RRDB} (Residual-in-Residual Dense Block). Como se muestran en la Figura \ref{fig:rrdb}, cada bloque está compuesto por cinco capas convolucionales, las conexiones densas están representadas por la concatenación de las salidas de diferentes capas. Esto quiere decir que cada capa tiene acceso a la información previa de una manera progresiva. Este tipo de arquitectura fue introducida para evitar el problema del desvanecimiento del gradiente durante el entrenamiento cuando se usa una red profunda \cite{gastineau2020residual}.
    \item El bloque de upsampling permite restaurar la resolución espacial de las imágenes después de todo el proceso dentro de los RRDB, para de esta manera tener una imagen en alta resolución.
    \item Los bloques de capas convulucionales al final de la red generadora permiten afinar el resultado del upsampling ya que son capaces de detectar las características espaciales de la imagen y a través de la convolución con filtros o kernels son capaces de re-ordenar los píxeles para generar una imagen de salida totalmente sintética que ingresará a la red discriminadora posteriormente.  
   \end{itemize}

\begin{figure}
 \centering
    \includegraphics[scale=0.7]{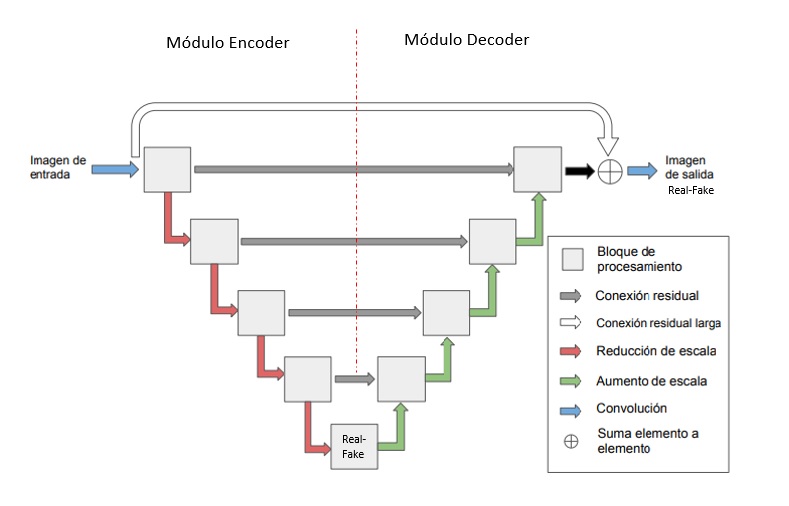}
    \caption{\label{fig:unet} Esquema de la arquitectura y procesos de la red discriminadora U-Net. Modificado de: González A. 2021} \cite{gonzalez2021metodos}. 
\end{figure}

Por su parte, para la red discriminadora fue utilizada una red \textit{U-Net} con normalización espectral. La normalización espectral es una técnica de normalización de los pesos que estabiliza el entrenamiento de la red discriminadora. Fue empleada para obtener un entrenamiento dinámico en cuanto a la supresión de errores visuales en las imágenes se refiere. La arquitectura basada en U-Net fue desarrollada en principio para la segmentación de imágenes biomédicas, adoptada como una red totalmente convolucional. Sin embargo, más tarde se adecuaría dentro del modelo de las GAN´s para actuar tanto como red generadora, como antagónica o discriminadora \cite{gonzalez2021metodos}. Su funcionamiento se basa en la concatenación de dos bloques, uno de encoder (izquierda) y otro de decoder (derecha) como se observa en la Figura \ref{fig:unet}. El bloque codificador progresivamente reduce la resolución de la entrada, capturando el contexto de la imagen global. El decodificador realiza un sobremuestreo progresivo, emparejando la resolución de salida a la de entrada y, por lo tanto, permite una localización precisa de las características.  Las conexiones residuales permiten enrutar datos de las  resoluciones coincidentes de los dos módulos, mejorando aún más la capacidad de la red para captar con precisión los detalles. El bloque de proceso central realiza una primera discriminación con la imagen reducida en su resolución. A la salida de la red, se vuelve a generar este procedimiento pero ya con la imagen re-escalada y ordenada a través de una operación de convolución \cite{schonfeld2020u}. La implementación de esta arquitectura para la red discriminadora permite proporcionar una retroalimentación detallada píxel por píxel al generador, mientras se mantiene la coherencia global de las imágenes sintetizadas, brindando también retroalimentación de la imagen global \cite{wang2021real}. 

\subsection{Reconocimiento de zonas de cultivo con CNN}
\subsubsection{Dataset}

Las imágenes fueron proporcionadas por el departamento de Agricultura de precisión del Ingenio Azucarero Valdéz. Son fotografías aéreas de los lotes de cultivo con mayor problemática en despoblación. Se han identificado en el departamento de experimentación agrícola, sustentado en resultados obtenidos por evaluaciones en campo. Posterior a ello, se han realizado vuelos de control y monitoreo generando así un dataset de 1200 imágenes, en las que se puede evidenciar zonas de siembra de caña y zonas despobladas.

\begin{figure}[H]
\centering
\subfloat[]{\includegraphics[width=0.4\linewidth]{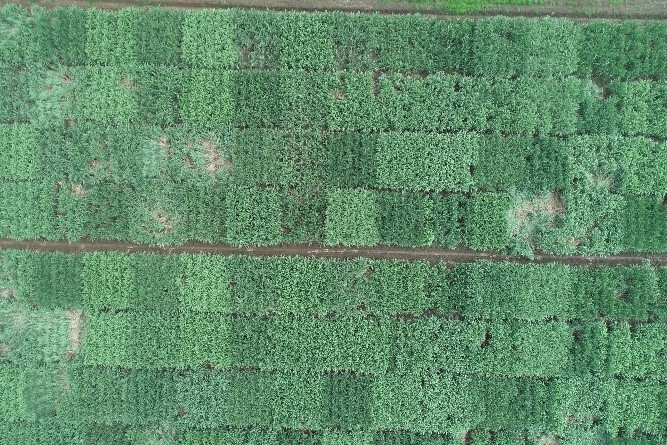}
\label{fig:cultivo}
}
\hfil
\subfloat[]{\includegraphics[width=0.4\linewidth]{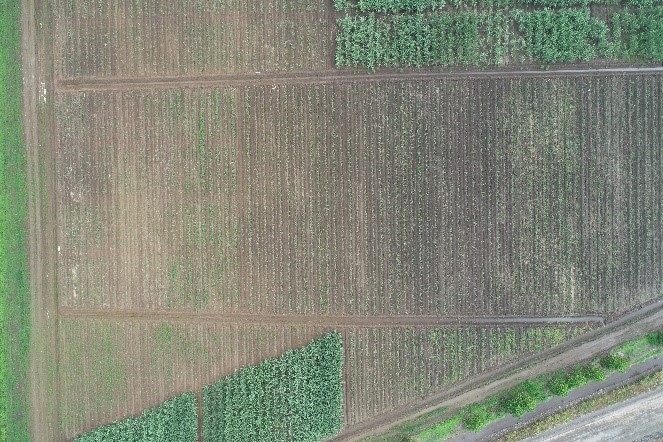}
\label{fig:nocultivo}
}
\caption{(a) Zona poblada de cultivo; (b) Zona despoblada de cultivo. Fuente: Compañía Azucarera Valdez S.A.}
\label{datasetingenio}
\end{figure}

La Figura \ref{datasetingenio} muestra ejemplos de fotografías aéreas de la plantación de Azucarera Valdez. En la fotografía a) se observa una zona de lote cultivado con diferentes variedades de caña para el entrenamiento del modelo y en la fotografía b) una zona con alta despoblación en la que la caña se encuentra en etapa de germinación, mismas que cuentan con una resolución horizontal y vertical de 72 ppp (puntos por pulgada), alto 3648 píxeles y ancho 5472 píxeles y presentan un formato JPG. Han sido organizadas en dos directorios: Zonas\_Pobladas (800 fotografías con un peso total de 6.5 GB) y Zonas \_Despobladas (800 fotografías con un peso de 6.8 GB). Al tratarse de un aprendizaje supervisado, se debe alimentar y entrenar la red definiendo entradas (Zonas cultivadas y Zonas Despobladas) y sus respectivas salidas (Clasificación de la categoría de la imagen y delimitación de las zonas de conflicto). Para la denominación de las imágenes se ha mantenido el nombre arrojado por el departamento de experimentación agrícola, en el caso de las fotografías obtenidas directamente desde el dron llevarán la codificación \textit{dji01} o el número de fotografía correspondiente, y en el caso de ser una ortofotografía procesada llevará la codificación ejemplo lote 004-034 debido a que luego de ser reconocidas y clasificadas estas serán devueltas a la empresa para obtener la ortofoto general del lote.



\subsubsection{Pre-procesamiento}
Esta etapa consiste en la mejora (si es necesario), conversión, re-dimensionamiento y normalización de las imágenes que constituyen nuestro dataset, de tal manera que tengan el formato adecuado para el entrenamiento del modelo de clasificación. Aquellas fotografías que por fallas de enfoque dejan de ser completamente nítidas, aumentamos su resolución con Real-ESRGAN. Debido a que la captura inicial de las fotografías es realizada con el programa \textit{Pix4D mapper}, éstas guardan información de georreferenciación, reflejando un peso de entre 7 y 8 MB lo que aumentaba el tiempo de entrenamiento y ejecución del código. Se tuvo que realizar un ajuste al tamaño de las imágenes del dataset por medio del programa \textit{Photoshop CS} para cambiar el tamaño a 300x300 píxeles y disminuir el peso en las imágenes. Finalmente, las imágenes son convertidas en arrays de datos numéricos que son normalizados mediante la librería \textit{NumPy}.

\subsubsection{División Train/Test }
Para la división del dataset en sus conjuntos de entrenamiento y test, partimos de la clasificación previa realizada de forma manual, donde se identificaron un total de 1600 imágenes que han sido dividas en dos sub-clasificaciones de 800 imágenes para identificación de siembra en caña y 800 para zonas donde se observa despoblación. Posteriormente fueron cargadas al código especificando dos conjuntos (Train, Test); el primero de entrenamiento (train) que contiene el 80 por ciento de los datos y el segundo de prueba (test) con el 20 por ciento de los datos restantes, a continuación se muestra la distribución de los datos.

\begin{table}[H]
\centering
\begin{tabular}{ccc}
\hline
\multicolumn{1}{|c|}{\textbf{Variable}} & \multicolumn{1}{c|}{\textbf{Número de Imágenes}} & \multicolumn{1}{c|}{\textbf{Porcentaje}} \\ \hline
\multicolumn{1}{|c|}{Dataset Original}      & \multicolumn{1}{c|}{1600}          & \multicolumn{1}{c|}{100\%} \\ \hline
\multicolumn{1}{|c|}{Dataset Z.Cultivadas}  & \multicolumn{1}{c|}{800}           & \multicolumn{1}{c|}{50\%}  \\ \hline
\multicolumn{1}{|c|}{Dataset Z.Despobladas} & \multicolumn{1}{c|}{800}           & \multicolumn{1}{c|}{50\%}  \\ \hline
                                            & \textbf{Dataset Zonas cultivadas}  &                            \\ \hline
\multicolumn{1}{|c|}{Dataset}               & \multicolumn{1}{c|}{800}           & \multicolumn{1}{c|}{100\%} \\ \hline
\multicolumn{1}{|c|}{Train Dataset}         & \multicolumn{1}{c|}{640}           & \multicolumn{1}{c|}{80\%}  \\ \hline
\multicolumn{1}{|c|}{Test Dataset}          & \multicolumn{1}{c|}{160}           & \multicolumn{1}{c|}{20\%}  \\ \hline
                                            & \textbf{Dataset Zonas Despobladas} &                            \\ \hline
\multicolumn{1}{|c|}{Dataset}               & \multicolumn{1}{c|}{800}           & \multicolumn{1}{c|}{100\%} \\ \hline
\multicolumn{1}{|c|}{Train Dataset}         & \multicolumn{1}{c|}{640}           & \multicolumn{1}{c|}{80\%}  \\ \hline
\multicolumn{1}{|c|}{Test Dataset}          & \multicolumn{1}{c|}{160}           & \multicolumn{1}{c|}{20\%}  \\ \hline
\end{tabular}
\vspace{0.5cm}
\caption{Subdivisión del dataset.}
\label{tab:DivisionDatasetCNN}
\end{table}
\subsubsection{Modelo}

Para la creación del modelo se usó Transfer learning, en este caso \textit{EfficientNet-B5}, debido a que es un modelo que se ajusta de manera más eficiente al equilibrar cuidadosamente la profundidad, el ancho y la resolución de la red, lo que conduce a un mejor rendimiento \cite{tan2019efficientnet}. EfficientNet-B5 ha sido entrenado con varias imágenes destacando lotes de agricultura, forestal y deforestación, el tamaño de entrada de las imágenes es 456*456*3, pero la cualidad de este modelo B5 radica en que las imágenes de entrada pueden diferir del tamaño original, para poder trabajar con imágenes más pequeñas y acelerar su procesamiento, diferenciándolo de sus versiones anteriores es uno de los modelos más actuales de la familia EfficientNet, tomando en cuenta de que estas características nos permiten entrenar de forma óptima al modelo lo que la familia de B6 a B7 generaba problemas de overfitting y en la familia de B0 a B4 se relacionaba problemas de underfitting \cite{ali2021multiclass}, eso convierte a EfficientNet-B5 en un modelo ideal para adaptarlo a nuestro problema de clasificación de imágenes. La Figura \ref{fig:CNNmodel} presenta un esquema de esta arquitectura: 

\begin{figure}[H]
 \centering
    \includegraphics[width=0.75\linewidth]{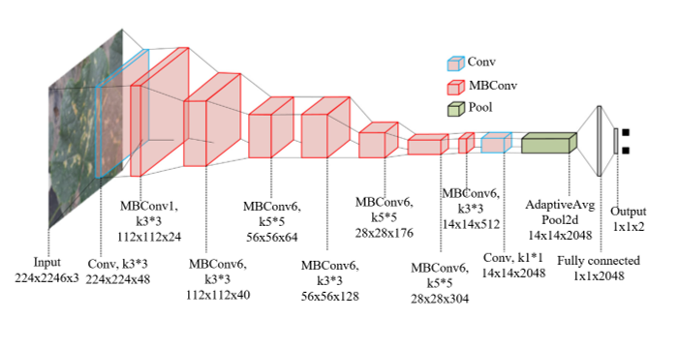}
    \caption{\label{fig:CNNmodel} Arquitectura de EfficientNet-B5. Tomado de: Zhang. P, 2020} \cite{zhang2020efficientnet}.
\end{figure}

Esta arquitectura se basa en una típica red neuronal convolucional (CNN), cuyo diseño toma en cuenta una capa de entrada con los valores de las imágenes de 224*224*3, la capa de entrada (Input) toma en cuenta el tamaño del lote, alto, ancho y tipo de canal que para el caso práctico será RGB conjuntamente con las probabilidades predeterminadas para cada clase \cite{cabrera2021entrenamiento}. Para adaptarlo a nuestro problema, es posible desconectar el clasificador del modelo predeterminado EfficientNet-B5 para obtener solamente la parte convolucional de extracción de sus características. El funcionamiento es el siguiente: la capa de entrada (Input) toma los valores predeterminados de las imágenes de entrada (224*224*3), posteriormente en las capas de convolución que en la arquitectura de EfficientNetB5 son 7 se toman grupos de píxeles cercanos de la imagen de entrada y se van operando matemáticamente contra una matriz pequeña llamada Kernel con un tamaño de (3*3*3) o (5*5*3), que recorre todas las neuronas de entrada y genera una nueva matriz de salida con la mitad de ancho y largo respecto a su valor original  (224*224 a 112*112) y así sucesivamente hasta tener un valor de salida de (14*14) este proceso denominado subsampling en el que reduciremos el tamaño de nuestras imágenes filtradas pero en donde deberán prevalecer las características más importantes que detectó cada filtro, donde el tipo más usado de subsampling es el Max-Pooling. Finalmente, mediante el uso de la capa AvgPool2D donde toma los valores promedios de los mapas de características donde esos valores pasarán al clasificador y mediante nuestra capa de salida se obtendrán las clases y gracias a la función "Softmax" se escogerá el valor mayor de la predicción.

\section{Experimentos}
\label{experiments}
En esta sección describimos la parte experimental del proyecto, donde se llevan a cabo los entrenamientos de los modelos de súper-resolución y clasificación utilizando los respectivos datasets. Utilizamos una plataforma computacional local y otra online debido a la capacidad de almacenaiento y memoria RAM que nos ofrece la plataforma \textit{Google Colaboratory}. Para el caso del modelo Real-ESRGAN, utilizamos una computadora de tipo portátil con un procesador Ryzen 5 5600x con 6 núcleos y 12 hilos a 3200 MHz, una memoria RAM de 32GB, GPU nVIDIA Geforce RTX 3060 de 12 GB de memoria y un disco duro de estado sólido de 1TB. Según la recomendación de los autores se utilizó el entorno de desarrollo de \textit{Visual Studio Code}, el cual es un editor de programación multi-plataforma desarrollado por Microsoft. Por su parte, para el entrenamiento del modelo convolucional caneCONV ID, se realizó en la plataforma on-line de hardware y software para aprendizaje automático de acceso gratuito que brinda 12 GB de RAM y 50 GB de almacenamiento en disco. Los códigos y comandos que se ejecuten mediante los notebooks de Colab usan GPU y TPU (Graphics Processing Unit y Tensor Processing Unit, respectivamente), que es un circuito integrado de aplicación específica y acelerador de Inteligencia Artificial, que no afecta el rendimiento de la computadora \cite{acosta2019inteligencia}.
 
\subsection{Entrenamiento}
\subsubsection{Real-ESRGAN}
Como ya se mencionó previamente, para los modelos en cuestión se aplicó la técnica de transfer learning. Siendo conceptualmente estrictos, para el modelo Real-ESRGAN se realizó un \textit{Fine-tuning} al modelo ya entrenado por los autores. El Fine-tuning es una técnica común para el aprendizaje por transferencia. Esta técnica resulta beneficiosa para entrenar nuevos algoritmos de aprendizaje profundo, cuando el conjunto de datos de un modelo existente y el nuevo modelo de aprendizaje profundo son similares entre sí. El fine tuning toma un modelo que ya ha sido entrenado para una tarea en particular y luego lo ajusta o modifica para que realice una segunda tarea similar \cite{Ekkis2019}. Para el entrenamiento de nuestro dataset primero se debió clonar el repositorio del modelo y almacenarlo dentro de una variable llamada \textit{Model1}, Posterior a ello, a través de la instrucción \textit{Model1.fit()} se procedió a inicializar el entrenamiento. Los hiper-parámetros establecidos para el entrenamiento del dataset son: un \textit{batch size} de 48, softmax como loss function, el optimizador \textit{Adam}, un número de 500 iteraciones o épocas y una tasa de aprendizaje de 0.0001.

\begin{figure}[H]
    \centering
    \includegraphics[width=1\textwidth]{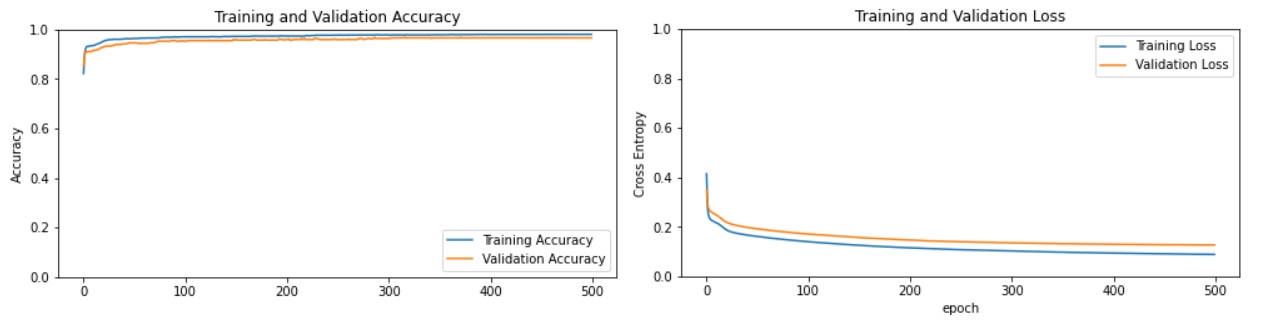}
    \caption{Curvas de aprendizaje para el modelo Real-ESRGAN. Se observa un muy buen ajuste en entrenamiento y validación. No son apreciables problemas de overfitting, ni underfitting demostrando que el modelo podrá tener un óptimo desempeño.}
    \label{fig:curvascnn}
\end{figure}

Este proceso tomó aproximadamente 34 horas continuas y la evolución del mismo se presenta en la Figura \ref{fig:curvascnn}, donde alcanzó su mejor precisión en la época 150 con un valor de precisión de 0.97 y un valor de pérdida de 0.12.

\subsubsection{CNN}
El uso de transfer-learning y fine-tuning nos ayuda a un entrenamiento más rápido del modelo sobre el dataset de entrenamiento el cual contiene 1280 imágenes, debido a que solo entrenamos las últimas 3 capas del modelo clasificador de EfficienteNet-B5. Los hiper-parámetros fueron: un batch size de 32, 5 épocas, con una función de perdida Softmax, el optimizador Adam, una tasa de aprendizaje de 0.001; según (Zhang. P, 2020) \cite{zhang2020efficientnet} recomienda que la tasa de aprendizaje debe ser menor a 0.01 para el uso de transfer learning.

\begin{figure}[H]
    \centering
    \includegraphics[width=1\textwidth]{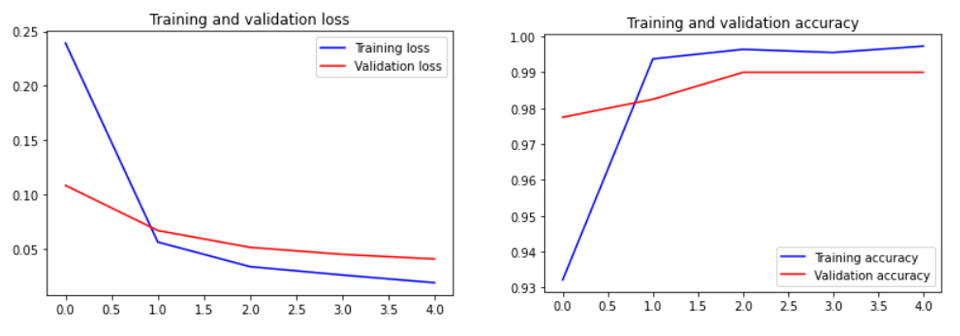}
    \caption{Curvas de aprendizaje de precisión y pérdida para los conjuntos de entrenamiento y validación.}
    \label{fig:fig8}
\end{figure}
 
Este procedimiento tomó aproximadamente 1 hora en ejecutarse completamente, la Figura \ref{fig:fig8} muestra que el entrenamiento alcanza altos valores de precisión desde la primera época con 0.9775, incrementando hasta llegar a un pico en la época 3 donde se estandariza con 0.9850 hasta las época 5, indicando que el ajuste entre el entrenamiento y la validación son correctos y no presentan problemas de overfitting ni underfitting, permitiendo así que el modelo sea adaptable al reconocimiento de problemas de despoblación y población en diferentes tipos de cultivos.

\subsection{Evaluación}
\subsubsection{Real-ESRGAN}
La mejora de la imagen o la mejora de la calidad visual de una imagen digital puede ser subjetiva. Decir que un método proporciona una imagen de mejor calidad puede variar de persona a persona. Por esta razón, es necesario establecer medidas cuantitativas empíricas para comparar los efectos de los algoritmos de mejora de imagen sobre la calidad de ésta \cite{gastineau2020residual}. El PSNR (Peak-signal to noise ratio) es una métrica para evaluar dicha característica, y es la relación entre la potencia máxima posible de una imagen, en cuanto a su señal y la potencia del ruido corruptor (distorsiones) que afectan la calidad de su representación \cite{Molinari2007}. Para estimar el PSNR de una imagen, es necesario comparar una imagen de baja calidad con su par de alta calidad a su máxima potencia posible.

La señal de una fotografía se refiere al detalle de representación de cada píxel que compone la imagen, mientras que el ruido es una alteración arbitraria de píxeles que no se corresponden con la luminancia y tonalidad real de la fotografía y que son apreciables a simple vista dado el tamaño que tienen \cite{Zafra2019}. Debido a que varias señales pueden tener un rango dinámico muy amplio, el PSNR es usualmente representado en términos de una escala de decibelios logarítmica \cite{clabaut2021model}. Los decibelios de manera general son una unidad que permiten expresar una relación entre dos potencias, en este caso la señal y el ruido de la imagen. El fundamento nos dice que mientras más alto sea el valor de PSNR, mayor será la calidad de la imagen reconstruida o re-escalada. Para el cálculo del valor de PSNR se utilizó la librería de \textit{OpenCvSharp}, a través de las siguientes instrucciones donde \textit{img1} representa la imagen de baja resolución e \textit{img2} representa su par de baja resolución:

\scriptsize
    \begin{verbatim}
                                               import cv2
                                               img1 = cv2.imread('img1.jpg')
                                               img2 = cv2.imread('img2.jpg')
                                               psnr = cv2.PSNR(img1, img2)
    \end{verbatim}
\normalsize

Los valores típicos del PSNR en la compresión de imagen y video con pérdida están entre 30 y 50 dB. Este parámetro estará en función de la profundidad de color de la imagen, misma que se refiere al número de bits necesarios para codificar y guardar la información de color de cada píxel en una imagen. El cálculo del PSNR se realizó para cada época durante el entrenamiento. Posteriormente se realizó un promedio para los valores de PSNR obtenidos durante la época 1 a 300 y entre la época 300 y 500 para tres períodos de entrenamiento obteniendo los resultados expuestos en la Tabla \ref{tab:promediosPSNR}.

\begin{table}[H]
\centering
\begin{tabular}{|c|c|c|c|}
\hline
\textbf{\begin{tabular}[c]{@{}c@{}}Periodo de\\ entrenamiento\end{tabular}} &
  \textbf{\begin{tabular}[c]{@{}c@{}}Promedio de PSNR\\  en dB para épocas\\  1-300\end{tabular}} &
  \textbf{\begin{tabular}[c]{@{}c@{}}Promedio de PSNR \\ en dB para épocas \\ 300-500\end{tabular}} &
  \textbf{\begin{tabular}[c]{@{}c@{}}Mejora de la \\ Imagen(re-\\ escalamiento)\%\end{tabular}} \\ \hline
1 & 30.91 & 30.00 & 0.39\%  \\ \hline
2 & 30.92 & 31.63 & 2.29\%  \\ \hline
3 & 33.69 & 34.72 & 15.72\% \\ \hline
\end{tabular}
\vspace{0.2cm}
\caption{Distribución promedios de PSNR durante el entrenamiento}
\label{tab:promediosPSNR}
\end{table}

Se puede observar que durante el tercer período de entrenamiento, a partir de la época 300 se logra alcanzar el valor más alto de PSNR (34.72 dB), por tanto, podríamos decir que es desde este punto donde se comienza a tener un mejor resultado en cuanto a la validación de la calidad de las imágenes usadas para el entrenamiento. De esta manera podríamos asociar un mejor resultado del entrenamiento después de al menos un período de entrenamiento inicial, sin embargo por recomendación de los autores, es necesario realizar al menos tres períodos de entrenamiento para observar una estabilización de los valores de PSNR. No obstante, generar más periodos de entrenamiento podría o no mejorar los resultados. 

\subsubsection{CNN}

\begin{figure}[H]
    \centering
    \includegraphics[width=0.25\textwidth]{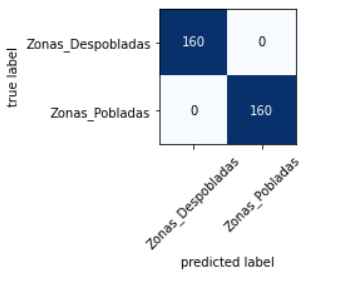}
    \caption{Matriz de confusión para evaluar el modelo de categorización de zonas de cultivo. Elaborado por: Autores.}
    \label{fig:fig9}
\end{figure}

Tomando en cuenta que el terreno no presenta texturas ideales en las fotografías aéreas, se ha delimitado netamente la zona de cultivo, y evaluado la mejor edad de la caña para realizar las pruebas (45 días a partir de cosecha) permitiendo observar las zonas realmente despobladas debido al crecimiento de la caña, obteniendo una respuesta positiva de aprendizaje. Las condiciones de evaluación son consideradas a partir de las imágenes de prueba (test), se define una variable que estandariza las dimensiones de la imagen y un modelo tipo categórico. Evaluamos mediante una matriz de confusión que es una herramienta útil para visualizar el desempeño del modelo en cuanto a la predicción con imágenes de test, mismas que previamente fueron divididas en las 2 categorías de Zonas pobladas y Zonas despobladas manualmente. Así nos ayuda a visualizar los aciertos y errores dentro del conjunto test, compara las imágenes predichas por el modelo contra las imágenes clasificadas manualmente, obteniendo que el modelo no se equivoca en predecir las dos categorías teniendo una total precisión (Figura \ref{fig:fig9}).

Definiendo de esta forma en la predicción con el 20\% del dataset utilizado y categorizado en test, la comparativa sobre la imagen original y el modelo de predicción para cada una de ellas (Figura \ref{fig:fig9}).

\begin{figure}[H]
    \centering
    \includegraphics[width=0.55\textwidth]{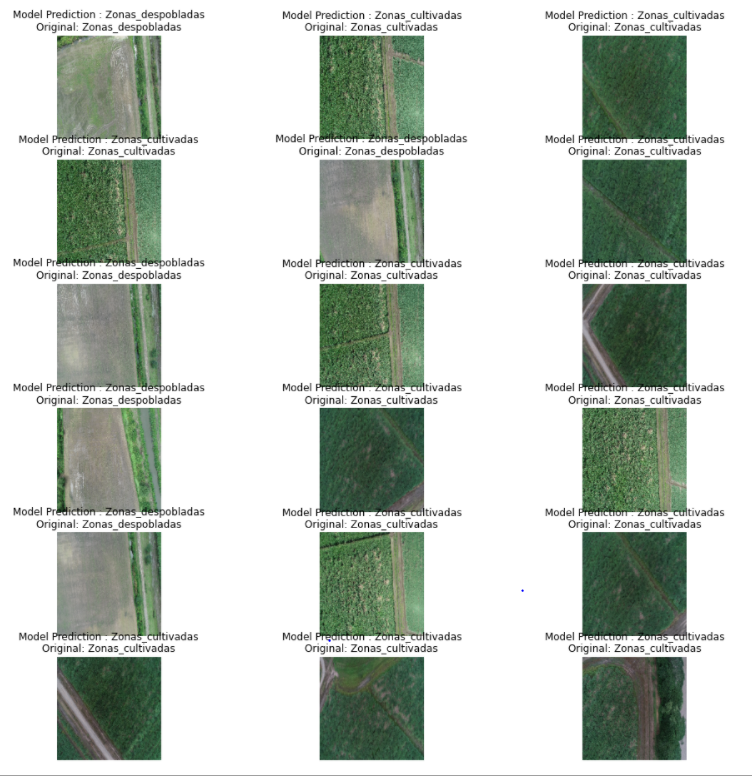}
    \caption{Comparación entre la clasificación original y la predicción del modelo. Se tiene una matriz de 3x6 en la que se han escogido imágenes aleatorias del dataset de test, y previamente categorizadas como zonas despobladas y zonas cultivadas, sobre la descripción de cada una encontramos la predicción del modelo en la cual podemos observar que la categorización del mismo tiene alta precisión con la realidad. Elaborado por: Autores.}
    \label{fig:fig99}
\end{figure}

\section{Resultados y discusión}
\begin{figure}[H]
    \centering
    \includegraphics[width=0.95\textwidth]{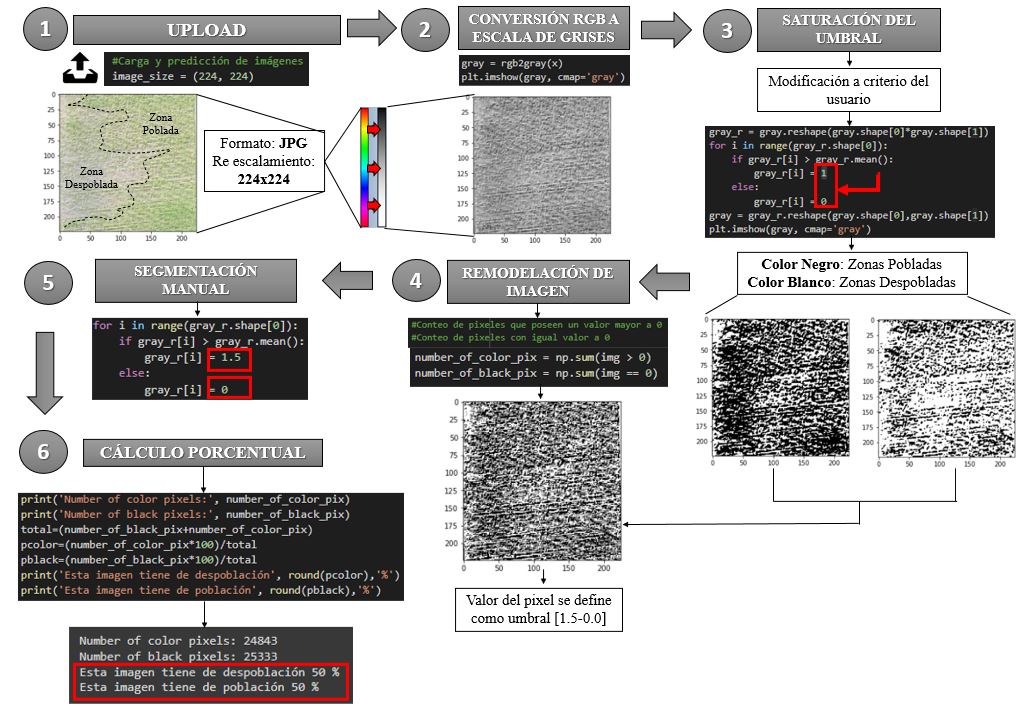}
    \caption{Diagrama ilustrativo para el cálculo de áreas despobladas. Elaborado por: Autores.}
    \label{fig:fig6}
\end{figure}

Para el cálculo de áreas cultivadas y áreas despobladas, y conocer sus respectivos porcentajes dentro de una fotografía aérea, una vez realizado el aumento de resolución y la clasificación de la imagen, seguimos el procedimiento ilustrado en la Figura \ref{fig:fig6}, apoyados en el uso de las librerías para el tratamiento de imágenes como \textit{Numpy}, \textit{CV2}, y \textit{Skimage}.

\begin{enumerate}
    \item Cargar la imagen previamente clasificada como zona despoblada, ya que al cargar una zona totalmente poblada el resultado sera de 100\% de población obteniendo un dato innecesario para la resolución del conflicto, las fotografías se obtienen en formato JPG reescaladas a un tamaño de 224x224 mediante el modelo de clasificación de fotografías con el uso de Efficientnet. 
    \item Dicha imagen es convertida a un arreglo de números modificada con el uso de la librería \textit{r.shape}, transformando la fotografía de 3 tres canales RGB a un solo canal en escala de grises.
    \item Realizamos una segmentación de la imagen mediante la librería de código abierto de \textit{matplotlib cmap}, variando la luminosidad de los colores representando el inicio y fin de los mismos, permitiendo así definir la saturación adecuada en la imagen, cuyo umbral podrá ser modificado a criterio del usuario.
    \item Una vez definida la saturación en la imagen y al ser capaces de aplicar el concepto de umbral global, gracias al submódulo \textit{SciPyndimage} se procede a remodelar la imagen tomando un valor de píxel y usándolo como umbral para distinguir los objetos del fondo, de los objetos del primer plano en una imagen.
     \item Segmentación manual: El valor del umbral puede especificarse entre 0-10 manualmente, si el valor del píxel es mayor que nuestro umbral, podemos decir que pertenece a un objeto, representando la región más luminosa (despoblación), caso contrario si su valor es menor se tratará como fondo, representando la región más oscura (población). Con ello la imagen se segmentará en dos grandes zonas: despoblación (región blanca) y población (región negra), sin embargo, hay limitaciones en este enfoque. Cuando no tenemos un contraste significativo en la escala de grises, o hay una superposición de los valores de píxeles, se vuelve muy difícil obtener precisión 
    \item Para obtener el cálculo final de despoblación en porcentaje, una vez obtenida la imagen el cálculo se ejecuta al contar todos los píxeles blancos y todos los negros y dividir cada clasificación para el total de los píxeles de la imagen obteniendo un valor porcentual.  
\end{enumerate}

\begin{figure}[H]
    \centering
    \includegraphics[width=0.5\linewidth]{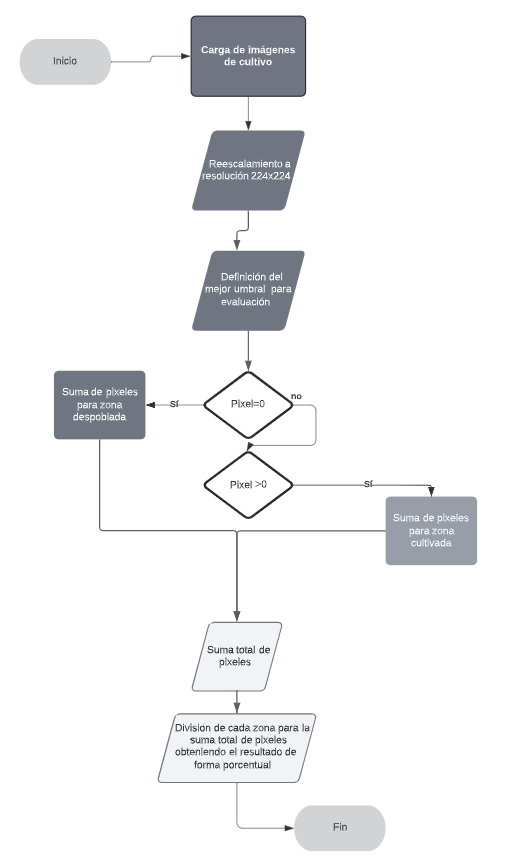}
    \caption{Diagrama de flujo para el cálculo porcentual de zonas pobladas y cultivadas con programación tradicional. Elaborado por: Autores.}
    \label{fig:fig10}
\end{figure}

A través del conteo de píxeles menores al umbral (población) y mayores al umbral (despoblación), se logra determinar la cantidad de píxeles que corresponden a cada región de la imagen. lo cual es posible gracias a la librería numpy que permitirá el cálculo y la suma de aquellos píxeles iguales a 0 tomados en cuenta como zonas cultivadas y mayores que 0 se sumarán en zonas despobladas, ésto haciendo referencia a píxeles con color. Para determinar el porcentaje de cada región con respecto al área total de la imagen, se divide la cantidad de los píxeles obtenidos para el número total de los mismos, con ello se logra obtener el porcentaje que ocupa cada región (población y despoblación) con respecto al área total (Figura \ref{fig:fig10}).

Los valores de precisión obtenidos para los modelos entrenados muestran excelente desempeño, por tanto es posible pasar a la etapa de predicción, donde se usaron imágenes fuera del dataset, siguiendo el orden del flujo de trabajo. En este sentido, fue importante tomar una sola imagen para procesarla con cada uno de los modelos propuestos en este trabajo.

Para fines prácticos, se procesó una imagen a la cual redujimos su resolución, para conocer la efectividad del modelo de súper-resolución. Además, para esta imagen la distribución de población y despoblación de cultivos ya era conocida por el departamento de agricultura de precisión de la compañía Valdez. Así se pudo tener un marco de referencia para corroborar la precisión de nuestros resultados. 

El código de inferencia para el modelo de Real-ESRGAN fue probado a través de la plataforma de Google Collaboratory, y fue llevado a cabo a través de 5 etapas. La primera etapa es referida a la importación de las librerías necesarias tales como son: BasicSR, GfpGan y facelibx, aquí también se descarga el modelo de Real-ESRGAN pre-entrenado. La segunda etapa comprende la subida o carga de las imágenes nuevas para el modelo, lo cual se lo realiza con la ayuda del módulo OS y la librería de Shutil. De esta manera podemos crear un \textit{widget} que nos permite insertar imágenes desde nuestro equipo para que estas puedan ser mejoradas. 

\begin{figure}[H]
    \centering
    \includegraphics[width=0.9\textwidth]{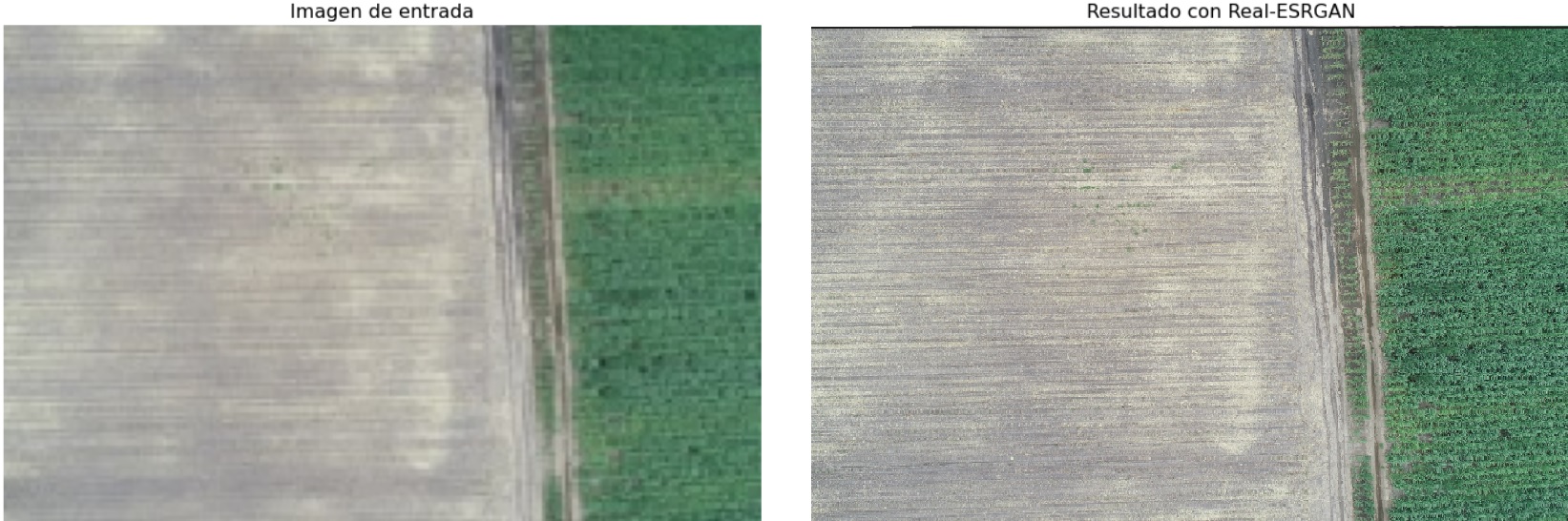}
    \caption{Comparación entre una imagen sin procesar vs. una imagen procesada con el modelo de Real-ESRGAN.}
    \label{fig:fig11}
\end{figure}

A continuación, en la tercera etapa ejecutamos el compilador a través de la siguiente línea de comando:

\begin{verbatim}
!python inferencerealesrgan.py -n RealESRGANx4plus -i upload --outscale 4 -half
\end{verbatim}

Donde: -n, --modelname: Nombre del modelo, -i, --input: dirección de la carpeta de la imagen cargada, --outscale: Factor de re-escalado de la imagen (este último factor puede ser expresado hasta un valor de 4, que es el valor de re-escalamiento máximo del modelo). En la siguiente etapa podemos tener una pre-visualización (Figura \ref{fig:fig11}). Finalmente, podremos descargar el resultado en formato ZIP dentro de nuestro equipo. 

\begin{figure}[H]
    \centering
    \includegraphics[width=0.4\textwidth]{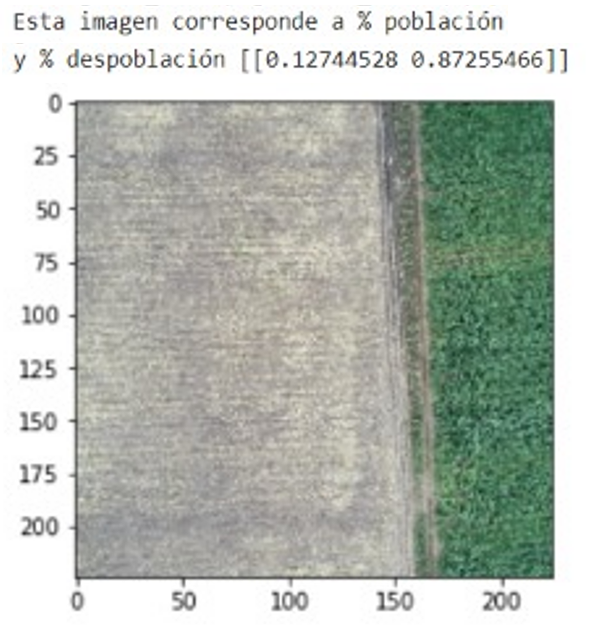}
    \caption{Predicción de una imagen de prueba fuera del dataset original. En la imagen se puede observar que el modelo ha predicho que un 12\% se ajusta a ser una imagen de zona cultivada mientras que un 87\% la ha reconocido como zona despoblada evidenciando así la precisión del modelo. Elaborado por: Autores.}
    \label{fig:fig12}
\end{figure}

Posteriormente, con la resolución de la imagen mejorada, es posible continuar con el siguiente paso dentro del flujo de trabajo. De esta forma la imagen optimizada entra al proceso de predicción de la zona poblada de cultivo y la despoblada de cultivo (Figura \ref{fig:fig12}), en el que se define el porcentaje de veracidad  que tiene el modelo para poder reconocer si es una zona cultivada o una zona despoblada. 

\begin{figure}[H]
    \centering
    \includegraphics[width=1\linewidth]{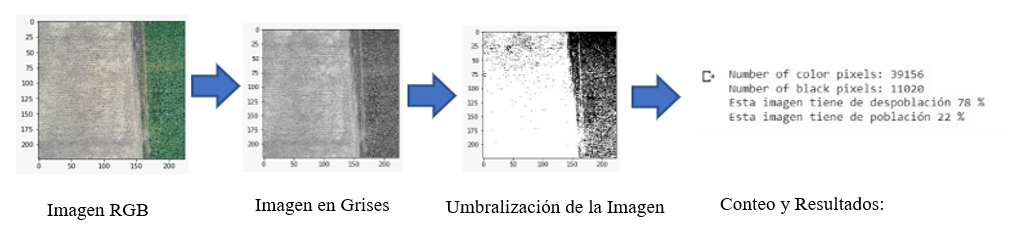}
    \caption{Proceso de umbralización de la imagen para el cálculo porcentual de superficies de población y despoblación de cultivos de caña. Elaborado por: Autores.}
    \label{fig:fig13}
\end{figure}

No obstante, se debe aclarar que el modelo arroja la probabilidad que la imagen pertenece a una Zona Despoblada y no son los porcentajes de cultivo o no cultivo, también se debe realizar una estandarización de parámetros con el objetivo de trabajar bajo los términos adecuados en cuanto al tamaño de la imagen. Se obtiene el porcentaje de precisión al reconocer zonas pobladas de zonas despobladas mediante el modelo de predicción (Figura \ref{fig:fig13}).

Finalmente, a través del código para el cálculo porcentual logramos determinar el porcentaje que le corresponde a las áreas pobladas y despobladas de cultivo reconocidas previamente, verificando de esta manera que nuestro flujo de trabajo trabaja de manera óptima con imágenes fuera del dataset y corroborando así los resultados obtenidos durante el entrenamiento.

\section{Conclusiones}

A través de la implementación de la Inteligencia Artificial y más concretamente las redes neuronales, es posible dar solución a problemáticas que actualmente pueden significar altos costos y tiempos de espera para el procesamiento de información. En nuestro caso hemos logrado diseñar un flujo de procesos destinados al tratamiento de imágenes aéreas, con la finalidad de calcular las áreas de población y despoblación de cultivos en los predios pertenecientes a la Compañía Azucarera Valdez S.A. 

Los modelos adoptados así como su arquitectura, fueron entrenados con los datasets creados por nosotros. De esta manera, aplicando las técnicas de transfer learning y Fine-tuning pudimos adaptarlos a nuestras necesidades de cara a enfrentar la problemática propuesta.

La utilización de las GANs, en el caso de Real-ESRGAN y las redes convolucionales para el modelo propuesto por nosotros, CaneCONV ID, permiten mejorar la resolución de imágenes afectadas por fallas en el proceso de captura de la fotografía e identificar superficies de población y despoblación de los cultivos de caña, respectivamente, evidenciando que el código utilizado durante todo el proceso funciona de forma óptima.

Para el proceso final de determinar el porcentaje de población y despoblación en las imágenes aéreas, se utiliza programación tradicional, de esta manera se puede lograr encontrar el umbral óptimo entre estas dos zonas de evaluación. Mediante el conteo de píxeles y acorde a la clasificación se tiene a las zonas cultivadas, con píxeles de máxima saturación tomando una coloración oscura y zonas despobladas con píxeles de mínima saturación tomando una coloración blanca y finalmente se definirá una relación en base al área total de la imagen para expresar cada parámetro porcentualmente.

Con los resultados obtenidos se determinó que los modelos son totalmente funcionales para el objetivo planteado. En primera instancia, la metodología adoptada podría adecuarse a otros problemas similares como el reconocimiento de malezas, plagas y enfermedades de cultivos, con el fin de optimizar parámetros en su análisis. 

El cálculo de despoblación con programación tradicional podría ser mejorado mediante el desarrollo de modelos a partir de redes neuronales que permitan definir el valor porcentual de forma automática. 

\section{Agradecimientos}

Deseamos extender nuestros agradecimientos a la Empresa Azucarera Valdez S.A., y al Ministerio de Ganadería y Agricultura, a través de su portal SIGTIERRAS, por proporcionarnos la información necesaria para la elaboración de nuestros datasets. De la misma manera al Ing. Christian Mejía, quien nos guió durante toda la elaboración del presente estudio.\\

%
%
\printbibliography
\end{document}